%% file: main.tex
\documentclass{article}

\usepackage{microtype}
\usepackage{graphicx}
\usepackage{subcaption}
\usepackage{booktabs} % for professional tables

\usepackage{hyperref}

\usepackage[preprint]{icml2026}

\usepackage{amssymb}
\usepackage{mathtools}
\usepackage{amsthm}
\usepackage{cutwin}
\usepackage{bm}
\usepackage{enumitem}
\usepackage{amsmath}
\usepackage{multirow}
\usepackage{color}
\usepackage{amsfonts}
\usepackage{nicefrac}
\usepackage{xcolor}
\usepackage{bbding}
\usepackage{pifont}
\usepackage{array} 
\usepackage{colortbl}
\usepackage{rotating}
\usepackage{adjustbox}
\usepackage{makecell}
\usepackage{fbox}
\usepackage{algpseudocode}
\usepackage{tabularx}
\usepackage{float}
\usepackage[most]{tcolorbox}
\tcbuselibrary{skins, breakable}
\tcbuselibrary{most}
\tcbset{fonttitle=\bfseries}

\definecolor{promptbar}{RGB}{200,220,245}
\definecolor{qcbar}{RGB}{210,210,210}

\tcbset{
  boxbase/.style={
    enhanced,
    breakable,
    colback=white,
    colframe=black!55,
    boxrule=0.5pt,
    arc=2pt,
    left=6pt,right=6pt,top=6pt,bottom=6pt,
    before skip=8pt, after skip=10pt,
    fonttitle=\bfseries,
    coltitle=black,
    colbacktitle=black!3,
    attach boxed title to top left={xshift=6pt,yshift=-2mm},
    boxed title style={
      boxrule=0.5pt,
      colframe=black!55,
      colback=black!3,
      arc=2pt,
      left=4pt,right=4pt,top=2pt,bottom=2pt,
    },
    fontupper=\footnotesize, % <-- add this (or \small)
  },
  promptbox/.style={
    boxbase,
    borderline west={2pt}{0pt}{promptbar},
  },
  qcbox/.style={
    boxbase,
    borderline west={2pt}{0pt}{qcbar},
  },
}

\definecolor{sgreen}{RGB}{30, 150, 30} 

\definecolor{cvprblue}{rgb}{0.21,0.49,0.74}
\definecolor{sgreen}{RGB}{30, 150, 30}
\definecolor{mycolor_blue}{HTML}{E7EFFA}
\definecolor{mycolor_green}{HTML}{E6F8E0}
\definecolor{mycolor_gray}{HTML}{ECECEC}
\definecolor{pearDark}{HTML}{2980B9}
\definecolor{textcolor1}{rgb}{0.25,0.5,0.5}
\definecolor{textcolor2}{rgb}{0.7,0.25,0.25}
\definecolor{linkc}{rgb}{0, 0.44, 0.74}
\definecolor{eqc}{rgb}{1, 0, 0}
\definecolor{myy}{RGB}{126,95,0}
\definecolor{mygray}{gray}{.9}
\definecolor{bblue}{RGB}{30,80,120}
\definecolor{mygray1}{gray}{.7}
\definecolor{ggray}{RGB}{127,127,127}
\definecolor{mygreen}{RGB}{93,174,86}
\definecolor{citecolor}{HTML}{229954}
\definecolor{light_green}{HTML}{F5FFFA}
\definecolor{LightCyan}{rgb}{0.88,1,1}
\definecolor{scolor}{RGB}{111,168,220}
\definecolor{hcolor}{RGB}{111,176,81}
\definecolor{ocolor}{RGB}{224,103,102}
\definecolor{wcolor}{RGB}{246,178,107}

\usepackage[capitalize,noabbrev]{cleveref}

\theoremstyle{plain}

\theoremstyle{definition}

\theoremstyle{remark}

\usepackage[textsize=tiny]{todonotes}

\icmltitlerunning{AIM-Bench: Benchmarking and Improving Affective Image Manipulation via Fine-Grained Hierarchical Control}

\begin{document}

\twocolumn[
  \icmltitle{AIM-Bench: Benchmarking and Improving Affective Image \\ Manipulation via Fine-Grained Hierarchical Control}

  \icmlsetsymbol{equal}{*}
  \icmlsetsymbol{lead}{$\dagger$}

  \begin{icmlauthorlist}
    \icmlauthor{Shi Chen}{xjtu,equal}
    \icmlauthor{Xuecheng Wu}{xjtu,equal,lead}
    \icmlauthor{Heli Sun}{xjtu}
    \icmlauthor{Yunyun Shi}{xjtu}
    \icmlauthor{Xinyi Yin}{zzu}
    \icmlauthor{Fengjian Xue}{xjtu}
    \icmlauthor{Jinheng Xie}{nus}
    \icmlauthor{Dingkang Yang}{fdu}
    \icmlauthor{Hao Wang}{xjtu}
    \icmlauthor{Junxiao Xue}{ZJL}
    \icmlauthor{Liang He}{xjtu}
  \end{icmlauthorlist}

  \icmlaffiliation{xjtu}{Xi'an Jiaotong University}
  \icmlaffiliation{zzu}{Zhengzhou University}
  \icmlaffiliation{nus}{National University of Singapore}
  \icmlaffiliation{fdu}{Fudan University}
  \icmlaffiliation{ZJL}{Zhejiang Lab}

  \icmlcorrespondingauthor{Heli Sun}{hlsun@xjtu.edu.cn}
  \icmlcorrespondingauthor{Hao Wang}{haowangx@xjtu.edu.cn}

  \icmlkeywords{Machine Learning, ICML}

  \vskip 0.24in
]

\printAffiliationsAndNotice{* Equal contribution; $\dagger$ Project lead}

\input{Sec/0_abs}

\begin{figure}[t!]
\centering
\includegraphics[width=\linewidth]{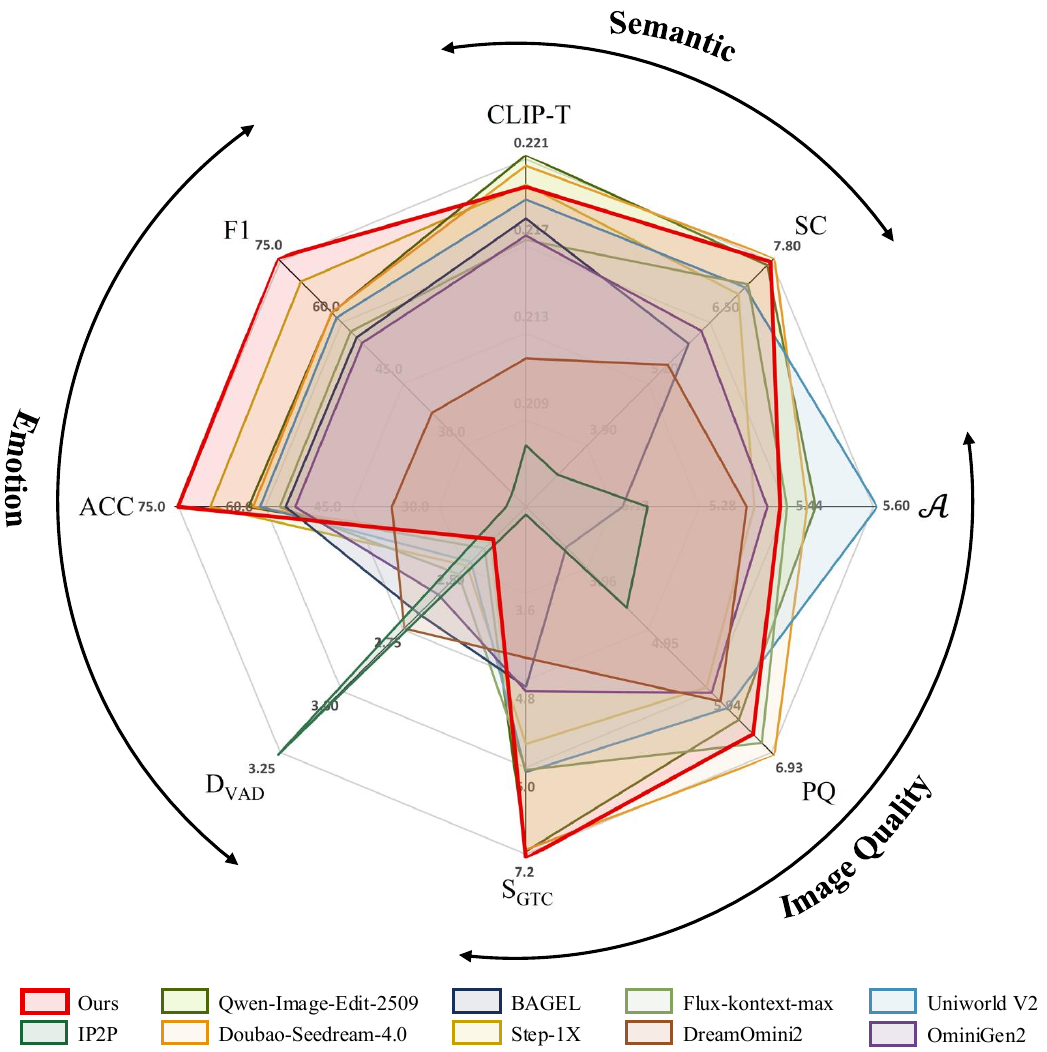}
\vspace{-1.30em}
\caption{Performance overview of expert editing models and unified large multimodal models on introduced AIM-Bench.}
\label{radar-fig}
\vspace{-1.30em}
\end{figure}

\begin{figure*}[t!]
\setlength{\belowcaptionskip}{-1.5em}
\centering
\includegraphics[width=\textwidth]{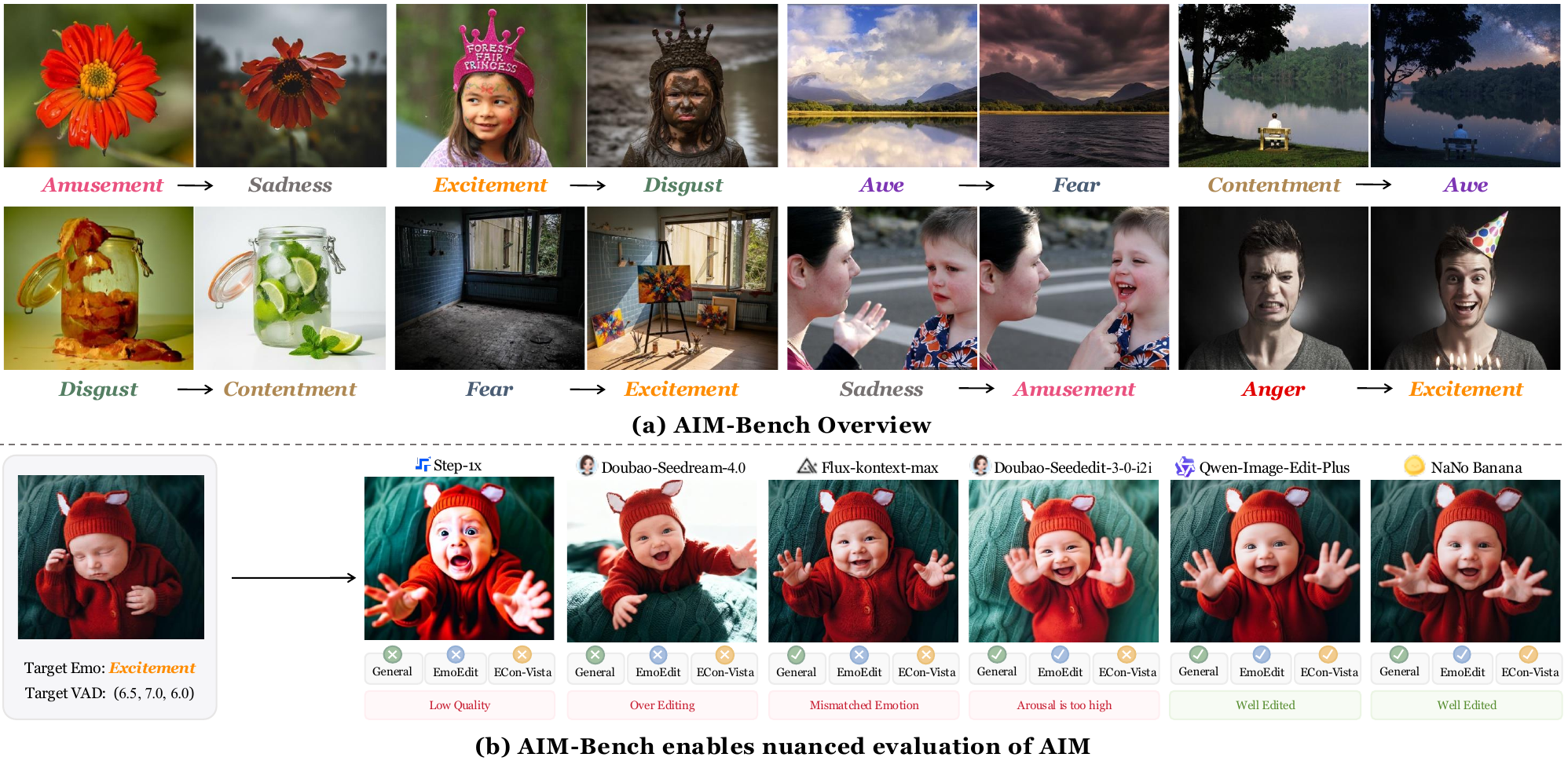}
\vspace{-1.0em}
\caption{The overview and motivation of proposed AIM-Bench. This benchmark comprises 800 high-quality samples, which cover 8 emotional categories and 5 image editing types.}
\label{fig-overall}
\vspace{0.5em}
\end{figure*}

\input{Sec/1_intro}

\begin{figure*}[t!]
\setlength{\belowcaptionskip}{-1.5em}
\centering
\includegraphics[width=\textwidth]{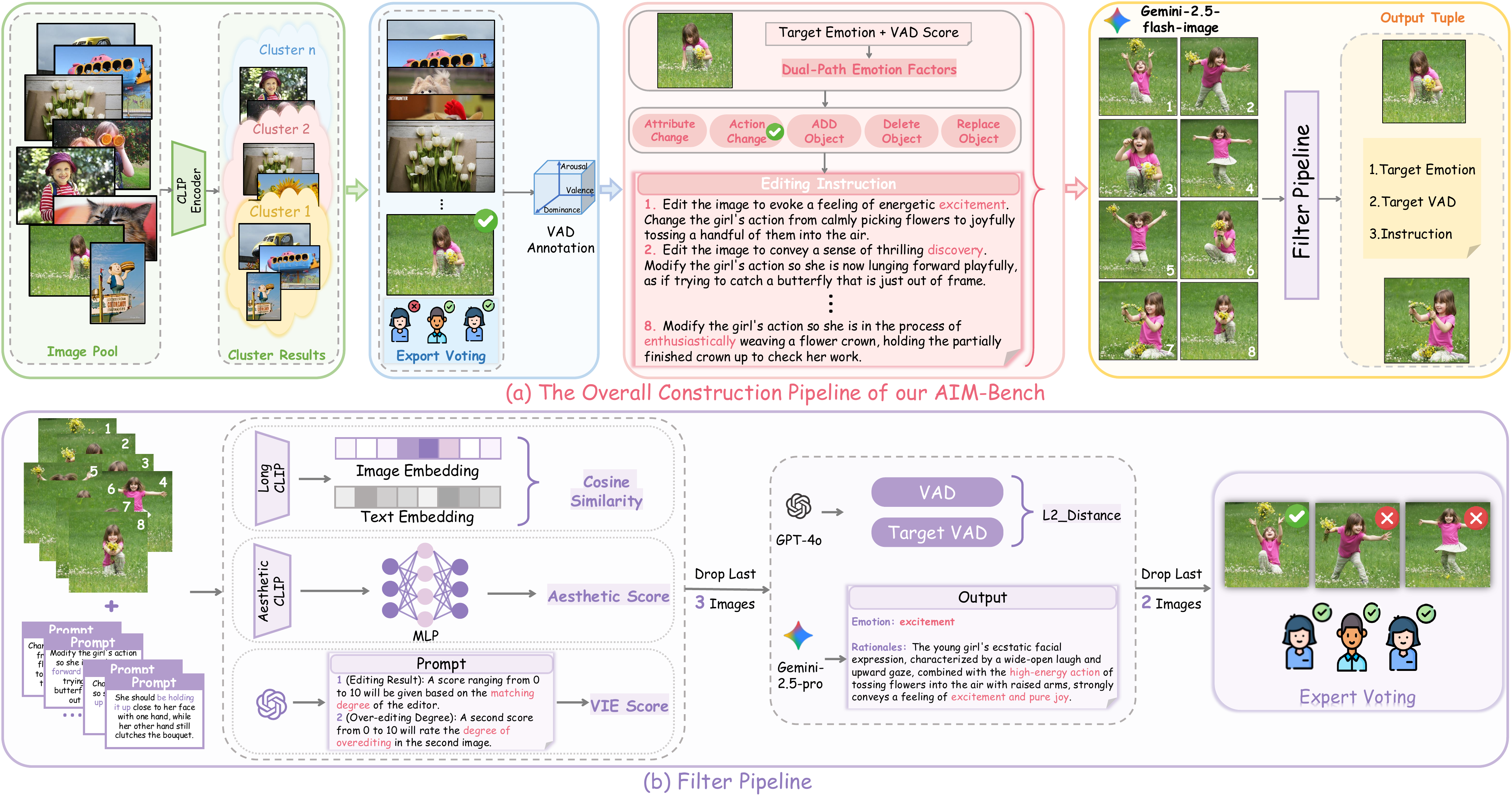}
\vspace{-1.3em}
\caption{\textbf{Overview of the AIM-Bench benchmark construction pipeline.} This pipeline consists of three main stages: \textbf{(1)} \textit{Data Collection}: clustering-based stratified sampling from EmoSet-118k to select 800 representative images; \textbf{(2)} \textit{Emotion-Targeted Image Editing}: generating emotion targets with VAD coordinates, creating diverse editing instructions, and producing candidate images; \textbf{(3)} \textit{Hierarchical Controlled Filtering}: progressively filtering candidates through quality assessment, affective alignment verification, and expert curation to output the final established benchmark.}
\vspace{0.5em}
\label{fig-pipeline}
\end{figure*}

\input{Sec/2_related}

\input{Sec/3_aimbench}
\input{Sec/4_aim40k}
\input{Sec/5_exp}
\input{Sec/6_conclusions}

\bibliographystyle{icml2026}
\bibliography{main}

\end{document}

%% file: Sec/0_abs.tex
\begin{abstract}
Affective Image Manipulation (AIM) aims to evoke specific emotions through targeted editing. Current image editing benchmarks primarily focus on object-level modifications in general scenarios, lacking the fine-grained granularity to capture affective dimensions. To bridge this gap, we introduce the first benchmark designed for AIM termed AIM-Bench. This benchmark is built upon a dual-path affective modeling scheme that integrates the Mikels emotion taxonomy with the Valence–Arousal–Dominance framework, enabling high-level semantic and fine-grained continuous manipulation. Through a hierarchical human-in-the-loop workflow, we finally curate 800 high-quality samples covering 8 emotional categories and 5 editing types. To effectively assess performance, we also design a composite evaluation suite combining rule-based and model-based metrics to holistically assess instruction consistency, aesthetics, and emotional expressiveness. Extensive evaluations reveal that current editing models face significant challenges, most notably a prevalent positivity bias, which stemming from inherent imbalances in training data distribution. To tackle this, we propose a scalable data engine utilizing an inverse repainting strategy to construct AIM-40k, a balanced instruction-tuning dataset comprising 40k samples. Concretely, we enhance raw affective images via generative redrawing to establish high-fidelity ground truths, and synthesize input images with divergent emotions and paired precise instructions. Fine-tuning a baseline model on AIM-40k yields a 9.15\% relative improvement in overall performance, demonstrating the effectiveness of our AIM-40k. Our data and related code will be made open soon.
\end{abstract}

%% file: Sec/1_intro.tex
\section{Introduction}
\label{sec:intro}

Affective images play a crucial role in online communication and web expression, serving as powerful tools to convey emotions and influence perceptions~\cite{wu2025avf,yang2025improving}, with broad applications in intelligent creation and emotion-aware AI systems~\cite{wu2025towards,xue2024affective}. This motivates affective image manipulation (AIM), which aims to evoke certain emotions through targeted image editing. Despite promising results, current image editing benchmarks~\cite{kawar2023imagic,imgedit} are primarily designed for general-purpose scenarios, lacking a comprehensive benchmark specifically designed for AIM. Meanwhile, current evaluations of model performance on AIM largely rely on generic image editing metrics or their customized criteria \cite{yang2025emoedit,aiedit}, which fail to capture the nuanced differences inherent in the complex concept of emotions. As displayed in Figure~\ref{fig-overall}, given the editing instruction \textit{Make the image evoke excitement}, existing benchmarks cannot distinguish between a moderately excited state and an extremely ecstatic one, nor can they quantify the precise degree of emotional transfer. This inability to differentiate emotional intensity severely limits fine-grained assessment and hinders continuous improvement in AIM research.

To tackle this gap, we introduce AIM-Bench, the first benchmark specifically designed to evaluate the performance of expert editing models and unified large multimodal models in AIM. First, we introduce the dual-path affective modeling scheme that integrates two solid psychological theories. Concretely, at the high-level semantic concept, we adopt the Mikels emotion model \cite{mikels2005emotional} to indicate the general direction of affective editing. At the fine-grained attribute concept, we employ the Valence-Arousal-Dominance (VAD) dimensional framework, utilizing the continuous manner to quantify subtle differences within each emotion.

Building upon this dual-path emotional modeling scheme, we then propose a hierarchically controlled workflow that primarily incorporates multi-stage human-in-the-loop collaboration for progressive data construction. Given an original image, we first choose an affective editing goal via our dual-path affective model, which comprises a discrete emotion label and corresponding VAD values. We then translate the affective goal into eight natural language instructions based on the most appropriate editing strategy to guide Gemini-2.5-flash-image~\cite{Gemini-2.5-flash-image} in generating candidate images. Subsequently, we select the optimal image that best embodies the predefined affective goal through a combination of automated model analysis and human expert review, which serves as the final reference target. In the end, this process leads to a complete visual editing tuple consisting of original image, editing goal, editing instructions, as well as target image.

Moreover, to enable systematic evaluation of affective image manipulation, we design a three-pronged evaluation suite that jointly assesses instruction-following fidelity, image aesthetics, and affective expressiveness. Based on our AIM-Bench, we perform a systematic evaluation of 13 advanced instruction-based image editing models. As illustrated in Figure~\ref{radar-fig}, which presents a holistic performance overview across semantic, aesthetic, and affective dimensions, we observe that current models exhibit varying trade-offs among these capabilities. We point out that a pervasive positivity bias emerges in these models. They tend to achieve higher accuracy and smaller VAD distances when editing for positive emotions than for negative ones. Besides, model capabilities exhibit a clear hierarchy from shallow attribute or color adjustments to holistic semantic-emotional reasoning.

To address this, we propose a scalable data engine utilizing an inverse repainting strategy to establish AIM-40k, a large-scale and balanced instruction-tuning dataset. Unlike a forward pipeline where models often fail to reach the target emotion, our inverse approach starts from images in EmoSet~\cite{yang2023emoset} that already possess accurate emotion labels. By synthesizing source images with divergent emotions while keeping the images in EmoSet~\cite{yang2023emoset} as targets, we ensure that the training process is guided by precise and reliable affective supervision. Crucially, before the inverse synthesis, we apply a generative repainting process to these targets in order to enhance their visual quality and fidelity. Experimental results show that baseline model fine-tuned on AIM-40k yields a 9.15\% relative performance gain, effectively mitigating the positivity bias and enhancing the emotional expressiveness of model.

Overall, the main contributions of this paper are three-fold:
\begin{itemize}
    \item We propose AIM-Bench, the first benchmark specifically designed for AIM, constructed via a dual-path affective modeling scheme and a hierarchical human-in-the-loop workflow. 
    \item We design a comprehensive evaluation suite that integrates rule-based and model-based metrics to holistically evaluate instruction-following fidelity, image quality, and emotional expressiveness, revealing critical insights such as the pervasive positivity bias in current models. 
    \item We propose a scalable data engine utilizing an inverse repainting strategy to construct AIM-40k, which is a large-scale and balanced instruction-tuning dataset for the AIM task.
\end{itemize}

%% file: Sec/2_related.tex
\section{Related Work}
\label{sec:Related Work}

\subsection{Text-Guided Image Editing}

The impressive advancement of diffusion models~\cite{croitoru2023diffusion,yang2023diffusion} has revolutionized image editing, establishing key methodologies ranging from instruction-based fine-tuning~\cite{brooks2023instructpix2pix} and cross-attention control~\cite{prompt-to-prompt} to explicit spatial conditioning~\cite{controlnet}. Building on these solid foundations, subsequent research emphasized the critical role of high-quality training datasets in enhancing edit fidelity~\cite{MagicBrush,hqedit,ultraedit}, while recent progress of unified large multimodal models (UMMs) have further elevated this popular field by enabling sophisticated semantic reasoning and contextual manipulations~\cite{MGIE,got}.

\subsection{AIM Evaluation}
Distinct from general-purpose editing, AIM focuses specifically on altering an image to evoke a target emotional response. Early explorations like AffectGAN \cite{affectgan} utilized Generative Adversarial Networks (GANs) to generate artistic images with designated emotional tones. EmoGen~\cite{yang2024emogen} introduced an emotion space to guide generation towards semantically clear and emotionally accurate outputs. EmoEdit \cite{yang2025emoedit} advanced the field further by constructing a large-scale dataset and a portable Emotion Adapter to enhance emotional impact by modifying image content. Echoing the trend in general-purpose editing, the latest methods like AIEdiT \cite{aiedit} have begun to utilize Multimodal Large Language Models (MLLMs) to enhance the model's capacity for emotional understanding.

\subsection{Image Editing Benchmarks}

The systematic evaluation of editing models has co-evolved alongside the generative technologies themselves. Initial benchmarks like TEdBench and EditBench \cite{editbench} established evaluation protocols focused on instruction fidelity and realism in general-purpose tasks. More recent and comprehensive benchmarks, such as I2EBench \cite{i2ebench} and ImgEdit-Bench \cite{imgedit}, have expanded this scope to cover a wider range of editing types and introduced more rigorous human evaluation protocols. However, a critical limitation pervades current landscape: these benchmarks were not designed for AIM. Their evaluation metrics fail to capture how an edit alters an image’s emotional impact. This gap makes it hard to assess the comprehensive AIM capabilities of advanced models.

%% file: Sec/3_aimbench.tex
\section{Benchmark Construction}
\label{sec:AIM-Bench}

\subsection{Benchmark Overview}

We introduce \textbf{AIM-Bench}, the first benchmark specifically designed for evaluating model performance on the AIM task. The AIM-Bench comprises 800 high-quality samples, covering 8 emotion categories and 5 editing types, ensuring comprehensive and fair evaluation. Figure~\ref{fig-pipeline} illustrates our benchmark construction workflow, which consists of three main stages: Data Collection, Emotion-Targeted Image Edit, and Hierarchical Controlled Filtering Pipeline.

\subsection{Data Collection}

We construct AIM-Bench based on EmoSet-118k \cite{yang2023emoset} to leverage its reliable human annotations. To ensure data diversity, we employ a clustering-based stratified sampling strategy.

\noindent \textbf{Visual Feature Clustering.}
We extract feature representations for all images using the CLIP-ViT-L/14 encoder \cite{radford2021learningtransferablevisualmodels} and apply MiniBatchKMeans clustering within each emotion category. The optimal cluster number $K^*$ is automatically determined by maximizing the silhouette coefficient based on the distances within the cluster and within the closest cluster.

\subsection{Emotion Oriented Image Editing}

Three trained experts independently selected high-quality images from the clusters, retaining only those with unanimous approval to ensure reliability. With this final set of 800 diverse source images, we now proceed to define appropriate editing objectives for each image and generate the corresponding natural language instructions to guide the overall process of image editing.

\noindent \textbf{VAD Annotation.} 
Fine-grained affective image manipulation requires continuous supervision in the VAD space. However, acquiring high-quality continuous VAD annotations manually at scale is prohibitively expensive and time-consuming. To address this, we utilize GPT-4o to extract VAD coordinates, as it demonstrates superior alignment with human perception in affective understanding. We rigorously validate the reliability of this automated annotation pipeline against human experts in the Appendix, exhibiting a high consistency that justifies its use for large-scale dataset construction.

\noindent \textbf{Emotion Target Generation.} 
We employ GPT-4o to perform complete VAD annotation on 2,000 images from EmoSet, constructing a dataset $\mathcal{D}_\text{VAD}$ containing 2,000 (emotion, VAD) pairs. For each original image $O_i$ with source emotion label $E_i^{\text{src}}$, we randomly sample a different emotion-VAD pair from this dataset as the editing target:
\begin{equation}
G_i = (E_i^{\text{tgt}}, \text{VAD}_i^{\text{tgt}}) \sim \mathcal{D}_\text{VAD},
\end{equation}
where  $E_i^{\text{tgt}} \neq E_i^{\text{src}}$. This design ensures that each affective goal is derived from real existing images, enhancing the authenticity and achievability of editing objectives.

\noindent \textbf{Instruction Generation.}
To translate abstract emotion goals into concrete visual manipulations, we design five core editing types based on the semantic distance and VAD differences: \textit{Attribute Change}, \textit{Action Change}, \textit{Replace Object}, \textit{Delete Object}, and \textit{Add Object}. Based on the original image, emotion goal, and these editing priors, we use Gemini-2.5-Pro to generate 8 distinct editing instructions for each image.

\noindent \textbf{Candidate Image Generation.}
We select Google's latest SOTA image editing model, namely Gemini-2.5-flash-image, to generate the candidate images. This process generates a candidate pool $\mathcal{C}_i = \{I_i^1, I_i^2, ..., I_i^8\}$ containing 8 edited variants for each original image.

\subsection{Hierarchical Controlled Filtering Pipeline}
\label{sec:ctlpip}

To select the optimal image $I_i^*$ that exhibits highest quality and best matches the affective editing goal from the candidates, we propose a multi-stage human-in-the-loop hierarchical filtering workflow, including three stages: Comprehensive Quality Assessment, Affective Alignment Verification, and Expert Curation and Final Decision. This progressively narrows down the 8 candidate images corresponding to each original image to the final 1 best image.

\subsubsection{Comprehensive Quality Assessment}

We evaluate the basic quality of candidate images from three complementary dimensions, which are detailed as follows:

\noindent \textbf{Semantic Consistency.}
We calculate the CLIP-T score to quantify the cross-modal semantic alignment between the generated image and the editing instruction. We employ LongCLIP~\cite{zhang2024long} for this task, as its capacity to support text sequences up to 248 tokens effectively prevents truncation, ensuring the complete semantic encoding of complex editing instructions.

\noindent \textbf{Aesthetic Quality.}
We employ the improved-aesthetic-predictor to quantify visual appeal. This model predicts a score $\mathcal{A}(I) \in [0, 10]$ aligned with human preferences using a CLIP-based MLP regressor.

\noindent \textbf{VIEScore.}
We employ VIEScore~\cite{ku-etal-2024-viescore} to evaluate the edited images across two dimensions: Semantic Consistency (SC) and Preservation Quality (PQ). Following the original protocol, we first compute the individual scores $S_{\text{SC}}$ and $S_{\text{PQ}}$ for a given triplet $(O, I, T)$. To provide a unified assessment, the final VIEScore is derived as the arithmetic mean of these two indicators:
\begin{equation}
\text{VIE}_\text{Score} = \frac{S_{\text{SC}} + S_{\text{PQ}}}{2}.
\end{equation}

\noindent \textbf{Weighted Quality Fusion.}
To output the results of a comprehensive assessment, we compute a weighted score, \textit{i.e.},
\begin{equation}
S_{\text{quality}} = 0.5 \cdot \text{VIE}_{\text{norm}} + 0.3 \cdot \text{CLIP-T}_{\text{norm}} + 0.2 \cdot \mathcal{A}_{\text{norm}},
\end{equation}
where $m_{\text{norm}}$ represents the score of metric $m$ after Min-Max normalization to the range $[0, 1]$.

Based on $S_{\text{quality}}$, we rank candidates in descending order and retain the top-5 images for affective assessment.

\subsubsection{Affective Alignment Verification}

We verify affective transformation effectiveness across discrete and continuous dimensions to filter the top-5 candidates down to the final 3 for human expert review.

\noindent \textbf{Discrete Emotion Classification.}
We employ Gemini-2.5-Pro~\cite{Gemini-2.5-Pro} to classify the generated images into the 8 Mikels emotion categories. This choice is motivated by the model's high accuracy of 72.0\% across 2,000 samples from the EmoSet validation set.This step validates whether the visual content successfully aligns with the target discrete emotion label $E^{\text{tgt}}$.

\noindent \textbf{Continuous VAD Alignment.}
To capture fine-grained affective nuances, we utilize GPT-4o to predict the image coordinates in the continuous VAD space. Let $\mathbf{p} = (V_p, A_p, D_p)$ and $\mathbf{t} = (V_t, A_t, D_t)$ denote the predicted and target VAD coordinates, respectively. We quantify the affective deviation using the Euclidean distance:
\begin{equation}
D_{\text{VAD}} = \|\mathbf{p} - \mathbf{t}\|_2.
\end{equation}

\noindent \textbf{VAD-Based Filtering.}
We implement a hierarchical filtering strategy where discrete emotion alignment serves as a prerequisite. Specifically, among the candidates that satisfy the discrete classification, we utilize the continuous VAD distance as the decisive metric for fine-grained ranking. We sort the candidates by $D_{\text{VAD}}$ and discard the bottom two variants with the largest deviations. These discarded candidates represent the poorest affective alignment, while the top-3 candidates are retained for the final human expert curation.

\subsection{Human Expert Curation}

The final stage employs a rigorous human expert review process to ensure the selected images meet the highest standards for quality, affective alignment, and other difficult-to-quantify aspects. We recruited three human experts who underwent standardized training on emotion theory and image quality assessment protocols.

For each set of top-3 candidate images, all three experts independently review and vote for the best image. The evaluation considers multiple aspects including affective consistency with the target emotion, visual quality, semantic coherence, and editing naturalness. The final image is determined by majority voting among the three experts.

If all the experts agree that none of the candidates adequately meet our requirements, the set is flagged for regeneration with the refined instructions. After completing the overall process, we output 800 high-quality triplets  $\{O_i, (E_i^{\text{target}}, VAD_i^{\text{target}}), T_i, I_i^{\text{edited}}\}_{i=1}^{800}$. The benchmark exhibits balanced distribution across emotion categories and editing types.

\section{Evaluation Suite}
\label{sec:evaluation}

Successful affective image manipulation requires a model to balance three fundamental capabilities: it must accurately execute the user's editing intent, maintain high visual realism despite potential drastic changes, and, most critically, effectively evoke the desired emotional response. Consequently, relying on a single metric is insufficient. To comprehensively assess model performance on AIM-Bench, we adopt a multi-dimensional evaluation suite integrating both rule-based and model-based metrics to measure \textbf{(i)} semantic fidelity to ensure instruction adherence, \textbf{(ii)} aesthetic quality to guarantee visual plausibility, and \textbf{(iii)} emotional expressiveness to verify the success of affective transfer.

\subsection{Semantic Fidelity}
This dimension evaluates whether a generated edit $I_g$ follows the text instruction. We report the CLIP-T metric computed by LongCLIP and the Semantic Consistency (SC) score from VIEScore, as defined in Section~\ref{sec:ctlpip}. In addition, we introduce a GPT-4o-based ground-truth consistency score, denoted as GTC, by comparing $I_g$ with the benchmark target $I_{gt}$ along two axes: semantic match ($S_{sm}$) and visual similarity ($S_{vs}$). Therefore, it can be formulated as:
\begin{equation}
S_{\text{GTC}} = \min(S_{sm}, S_{vs}).
\end{equation}

\subsection{Aesthetic Quality}
This dimension assesses the overall visual appeal and the absence of artifacts. We compute the aesthetic score $\mathcal{A}(I_g)$ using the improved-aesthetic-predictor, and use the PQ score from VIEScore to quantify structural fidelity and unnatural distortions.

\subsection{Emotional Expressiveness}
This dimension evaluates the alignment between $I_g$ and target affective goals. For discrete emotion classification, we employ Gemini-2.5-Pro~\cite{Gemini-2.5-Pro} to predict labels and report Accuracy (ACC) and F1-score. The accuracy can be  formulated as:

\begin{equation}
\text{ACC} = \frac{1}{N} \sum_{i=1}^{N} \mathbb{I}(\hat{y}_i = y_i),
\end{equation}
where $N$ is the sample size, and $\hat{y}_i, y_i$ denote the predicted and ground-truth emotion labels for the $i$-th image, respectively. $\mathbb{I}(\cdot)$ is the indicator function.

For continuous affect, we report the mean $D_{\text{VAD}}$ over all test samples, where $D_{\text{VAD}}$ is defined in Section~\ref{sec:ctlpip}.

\noindent \textbf{Overall Score Aggregation.}
We normalize all metrics to the $[0,1]$ range using fixed upper and lower bounds derived from empirical experimental data and average them to obtain a unified overall score.

%% file: Sec/4_aim40k.tex
\section{AIM-40k Construction}
\label{sec:aim40k}

\begin{table*}[t!]
\centering
\caption{The performance evaluations of advanced models in our metrics~(\%) on AIM. Note that we highlight the best performance in \textbf{bold} and \underline{underline} the second performance. The ``Optimization Analysis'' section highlights the improvements of our method compared to the baseline.}
\vspace{-0.4em}
\setlength{\arrayrulewidth}{0.40pt}
\renewcommand\arraystretch{1.1} 
\resizebox{\textwidth}{!}{
\begin{tabular}{lcccccccccc}
\toprule[0.40pt]
\multirow{2}{*}{Method} & \multirow{2}{*}{Organization} & \multicolumn{2}{c}{Semantic} & \multicolumn{3}{c}{Image Quality} & \multicolumn{3}{c}{Emotion} & \multirow{2}{*}{Overall~$\uparrow$}\\

\cmidrule(lr){3-4} \cmidrule(lr){5-7} \cmidrule(lr){8-10}
&&  CLIP-T~$\uparrow$ & SC~$\uparrow$  & $\mathcal{A}$~$\uparrow$  & PQ~$\uparrow$  & $S_{\text{GTC}}$~$\uparrow$   & ACC~$\uparrow$ & F1~$\uparrow$ & $D_{\text{VAD}}$~$\downarrow$    \\ 
\midrule

\multicolumn{11}{l}{\textit{\textcolor{purple!80}{\textbf{Commercial proprietary models}}}} \\
Doubao-Seedream-4.0~\cite{doubao_seedream_4_0}  & ByteDance      & \underline{0.2211} & \textbf{8.28} & 5.53 & \textbf{6.97} & 7.06 & 61.55 & 61.49 & \underline{2.54} & \underline{48.94} \\
Qwen-Image-Edit-Plus~\cite{qwenimagetechnicalreport} &  Alibaba     & 0.2173 & 7.74 & 5.52 & 6.11 & 6.41 & 58.39 & 58.17 & 2.72 & 40.25  \\
Flux-kontext-max~\cite{flux} & Black Forest Labs  & 0.2176 & 7.72 & 5.49 & \underline{6.77} & 5.98 & 57.01 & 56.92 & 2.67 & 41.14  \\
Doubao-Seededit-3-0-i2i~\cite{doubao_seededit_3_0} & ByteDance          & 0.2175 & 7.85 & 5.42 & 6.02 & 6.03 & 52.73 & 52.63 & 2.65 & 38.79 \\
Flux-kontext-pro~\cite{flux} & Black Forest Labs      & 0.2159 & 7.52 & 5.45 & \underline{6.77} & 5.70 & 54.77 & 54.03 & 2.73 & 38.25  \\
\midrule

\multicolumn{11}{l}{\textit{\textcolor{purple!80}{\textbf{Open-source Models}}}} \\
Flux-kontext-dev~\cite{flux} & Black Forest Labs      & 0.2173 & 7.14 & \underline{5.55} & 5.66 & 5.58 & 50.88 & 50.65 & 2.77  & 33.12 \\
Uniworld V2~\cite{uniworld} & PKU    & 0.2195 & 7.66 & \textbf{5.66} & 6.22 & 6.01 & 60.38 & 60.27 & 2.62 & 41.57 \\
Step-1X~\cite{step1x-edit} & StepFun    & 0.2202 & 7.52 & 5.43 & 5.89 & 5.63 & \underline{68.84} & \underline{68.87} & 2.63  & 41.17 \\
DreamOmni2~\cite{dreamomni2} & CUHK   & 0.2120 & 6.02 & 5.42 &  6.11 & 4.45  & 37.88 & 37.47 & 2.88 & 24.47 \\
BAGEL~\cite{bagel} & ByteDance & 0.2186 & 6.47 & 5.19 & 3.66 & 4.85  & 56.00 & 55.52 & 2.83 & 28.59 \\
OmniGen2 \cite{omnigen2} & BAAI & 0.2178 & 6.74 & 5.46 & 5.97 & 4.91 & 54.37 & 54.19 & 2.75 & 32.53 \\
IP2P \cite{brooks2023instructpix2pix}      & UCB & 0.2079 & 3.69 & 5.23 & 4.62 & 2.51 & 18.38 & 18.26 & 3.39 & 12.11\\
\midrule
Qwen-Image-Edit-2509~\cite{qwenimagetechnicalreport} & Alibaba      & \textbf{0.2216} & 8.13 & 5.54 & 6.40 & \underline{7.09} & 62.34 & 61.44 & 2.57 & 47.09 \\
\rowcolor{cvprblue!20}
Qwen-Image-Edit-2509-LoRA~(\textbf{Ours}) & \textbf{--} & 0.2201 & \underline{8.20} & 5.48 & 6.63 & \textbf{7.18} & \textbf{74.37} & \textbf{74.51} & \textbf{2.53} & \textbf{51.40} \\
\hline
\end{tabular}
}
\vspace{-1em}
\label{tab:main-eval_maze}
\end{table*}

\begin{figure}[t!]
\centering
\includegraphics[width=\linewidth]{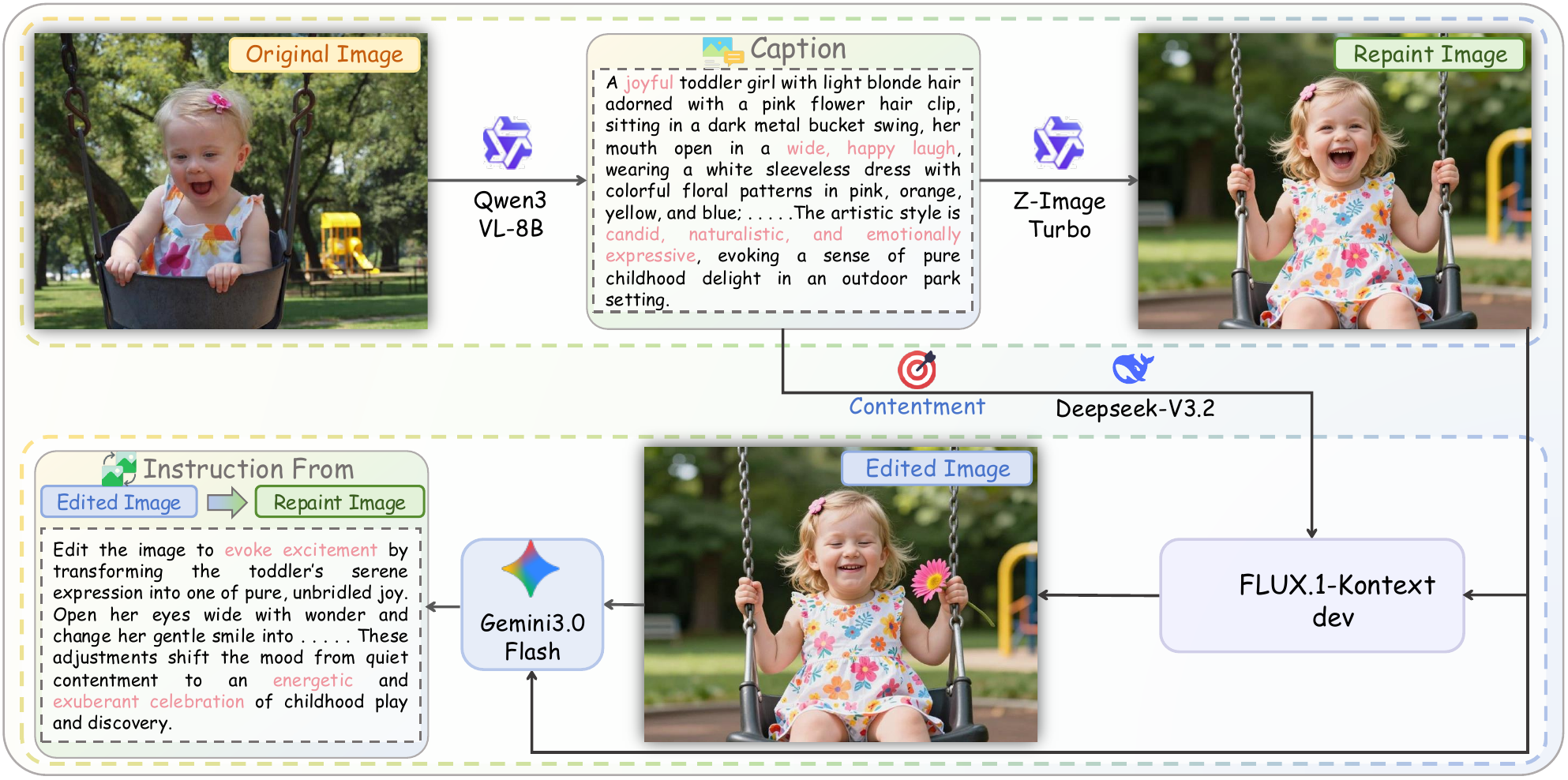}
\caption{The construction pipeline of AIM-40k. We introduce a rigorous repainting and inverse editing workflow to ensure both visual fidelity and affective alignment.}
\vspace{-1.8em}
\label{fig:pipeline}
\end{figure}

Existing affective image editing datasets often face the issue of low source image quality, and current image editing models frequently struggle to accurately evoke the intended target emotions. To address these challenges, we introduce \textbf{AIM-40k}, a large-scale AIM dataset constructed via a novel Inverse Repainting Strategy. As illustrated in Figure~\ref{fig:pipeline}, unlike traditional forward pipelines, our approach adopts a repainting and inverse editing workflow to establish high-quality ground truths and synthesize divergent inputs.

\subsection{High-Fidelity Ground Truth via Repainting}
\label{sec:repainting}

The first stage aims to establish visual ground truth images exceeding the quality of original source images while retaining validated emotional semantics. Directly using original images from EmoSet as training targets limits the upper bound of model performance due to compression artifacts.

\noindent \textbf{Repainting-Based Restoration.}
We employ Qwen3-VL-8B~\cite{qwen3technicalreport} to generate dense and descriptive captions for images sampled from EmoSet. Subsequently, we utilize Z-Image-Turbo to generate a high-fidelity Repaint Image conditioned on these captions. This step ensures the model targets are artifact-free and aesthetically pleasing.

\noindent \textbf{Semantic Consistency Verification.}
To guarantee that the quality enhancement does not alter the original semantic content, we calculate the cosine similarity between the original and repainted images using a CLIP encoder. We enforce a strict threshold where CLIP-I scores must exceed 0.85 to filter out deviant samples, yielding a robust set of verified ground truth targets.

\subsection{Inverse Instruction Tuning Data Synthesis}
\label{sec:inverse_synthesis}

A core challenge in constructing AIM datasets is the accuracy of the target emotion. We resolve this via an Inverse Synthesis approach where we treat the verified Repaint Image (denoted as $I_{gt}$) as the destination and the edited image as the input, ensuring the target emotion is grounded in the validated dataset distribution.

\noindent \textbf{Divergent Input Synthesis.}
To synthesize a source image $I_{input}$ with a divergent emotion, we first employ DeepSeek-V3.2~\cite{deepseekai2025deepseekv32pushingfrontieropen} to generate an emotion-shifting prompt based on the caption of $I_{gt}$. This prompt, combined with $I_{gt}$, is processed by FLUX.1-Kontext dev to synthesize the input image $I_{input}$, which serves as the challenging starting point for the editing task.

\noindent \textbf{Instruction Generation.}
To obtain the final training instruction guiding the forward transformation ($I_{input} \rightarrow I_{gt}$), we utilize Gemini-3.0-Flash~\cite{Gemini-3-flash} to analyze the visual differences between the pair. It generates a precise natural language command that explicitly maps the semantic shift from the synthesized input back to the ground truth, completing the data construction.

AIM-40k comprises 40,000 triplets ($I_{input}$ -- Instruction -- $I_{gt}$), evenly distributed across 8 emotion categories. By combining high-fidelity ground truths with synthesized divergent inputs, the dataset guarantees precise affective supervision and abundant semantic variations, establishing a robust foundation for advanced AIM research.

%% file: Sec/5_exp.tex
\section{Experiments}
\label{sec:exps}

\subsection{Experimental Setup}

We select 13 advanced instruction-based image editing models for comprehensive evaluation, categorized into official APIs and self-deployed open-source models. For all models, we strictly adopt the official default parameter settings provided by the developers to ensure a fair assessment of their intended performance. 

For our method, we initialize the backbone with Qwen-Image-Edit-2509~\cite{qwenimagetechnicalreport}. We inject LoRA layers with a rank of 32 into the diffusion transformer (DiT) modules. Training is performed on eight NVIDIA A100 GPUs using the DiffSynth-Studio framework. We optimize the model using the AdamW optimizer at a fixed learning rate of $1 \times 10^{-4}$. The fine-tuning process spans 1 epoch on the AIM-40k dataset, taking approximately 10 hours.

\begin{figure*}[t!]
\setlength{\belowcaptionskip}{-1.5em}
\centering
\includegraphics[width=\textwidth]{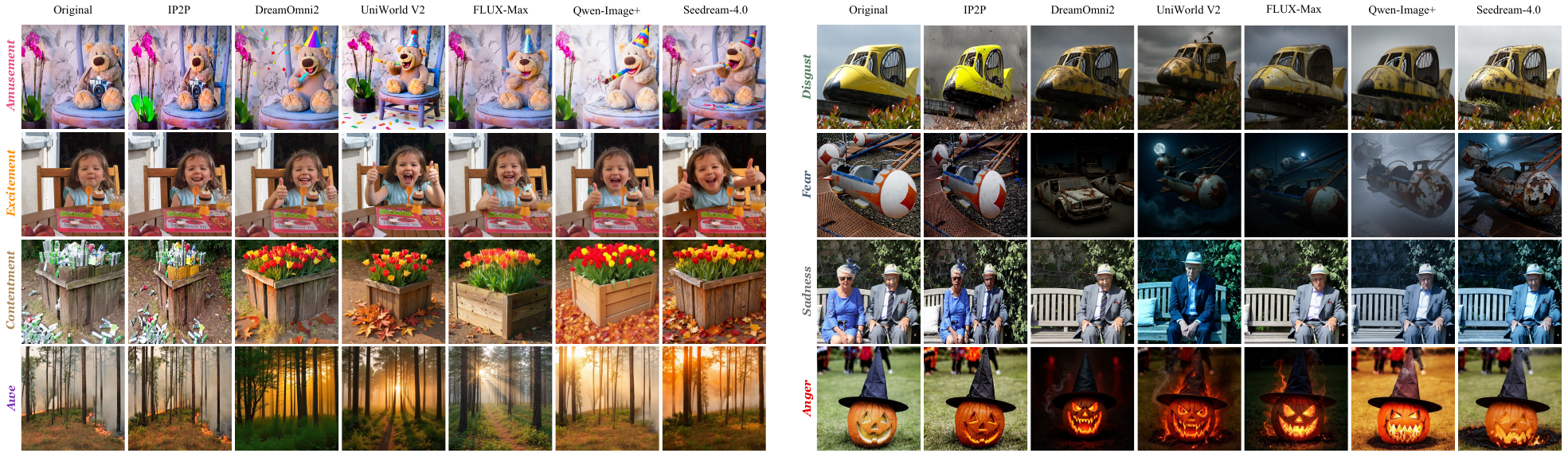}
\vspace{-1.3em}
\caption{Qualitative comparison of emotion editing across representative models. We visualize eight emotion categories (Rows 1-4: positive emotions, \textit{i.e.}, Amusement, Awe, Contentment, Excitement. Rows 5-8: negative emotions, \textit{i.e.}, Disgust, Sadness, Fear, Anger).}
\vspace{0.5em}
\label{fig-case-study}
\end{figure*}

\subsection{Main Results}

Our comprehensive evaluation across 13 leading instruction-based image editing models reveals significant insights into their capabilities, particularly in the nuanced domain of emotion editing. The main performance metrics are summarized in Table~\ref{tab:main-eval_maze}. To provide a deeper understanding of the emotion-related scores (ACC, $D_{\text{VAD}}$), we conduct a detailed breakdown of model performance on positive and negative emotion categories, as presented in Table~\ref{tab:emotion_breakdown} below.

\noindent\textbf{Overall Performance Leaders.}
Among the baseline models, the open-source model Qwen-Image-Edit-2509~\cite{qwenimagetechnicalreport} demonstrates strong capabilities, particularly in positive emotion editing (\textit{i.e.}, 73.87\% ACC). However, its performance significantly declines when handling negative emotions (\textit{i.e.}, 50.76\% ACC), highlighting a critical imbalance in current state-of-the-art methods. Similarly, proprietary models such as Doubao-Seedream-4.0 exhibit competitive results but still maintain a noticeable performance disparity between positive and negative editing tasks.

\noindent\textbf{The Positivity Bias in Baseline Models.}
A pervasive trend observed across the majority of baseline models is a significant performance gap between generating positive and negative emotions. Most models achieve higher accuracy and superior VAD distance scores when editing for positive emotions compared to negative ones. For instance, the Qwen-Image-Edit-2509~\cite{qwenimagetechnicalreport} model exhibits a dramatic drop of over 23 percentage points in accuracy from positive (73.87\%) to negative (50.76\%) tasks. This positivity bias suggests that current models are substantially more proficient at synthesizing pleasant emotional scenes, likely due to the prevalence of positive imagery in their pre-training datasets.

\noindent\textbf{Performance Improvements with AIM-40k.}
To elevate model performance and address the baseline imbalance, we fine-tuned Qwen-Image-Edit-2509~\cite{qwenimagetechnicalreport} on AIM-40k. As summarized in Table~\ref{tab:main-eval_maze}, our model consistently outperforms the baseline, achieving a significant \textbf{9.15\%} improvement in the overall score. This gain stems from joint enhancements in semantic fidelity, aesthetic quality, and emotional expressiveness, indicating that affective supervision does not come at the expense of instruction following or visual realism. Notably, our method achieves SOTA performance on emotion metrics while simultaneously boosting image quality, effectively verifying the rationality and effectiveness of the AIM-40k dataset construction.

\begin{figure}[t]
\centering
\includegraphics[width=\linewidth]{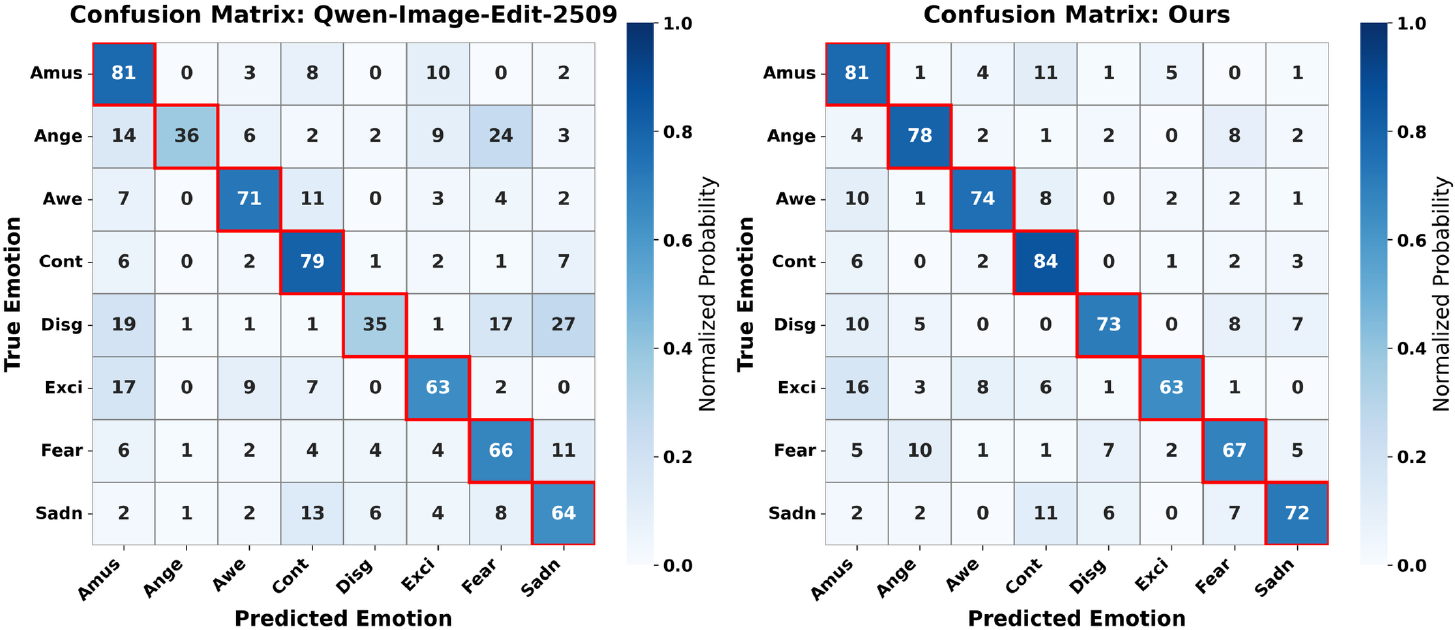}
\caption{AIM-40k confusion matrices: Unlike the baseline (Left) which misclassifies negative emotions, our model (Right) shows a sharper diagonal, effectively mitigating positivity bias.}
\vspace{-2.0em}
\label{fig:confusion_matrix}
\end{figure}

\noindent\textbf{Fine-grained Performance Analysis.}
To investigate these improvements, we analyze the distribution in Table~\ref{tab:emotion_breakdown} and Figure~\ref{fig:confusion_matrix}. The baseline exhibits a clear positivity bias: while strong on positive emotions, it suffers from substantial off-diagonal confusion among negative categories, collapsing distinct negative states into ambiguous clusters. In contrast, our tuned baseline model displays a pronounced diagonal concentration across all classes. With diverse and balanced supervision, AIM-40k enables learning discriminative cues, boosting negative-emotion accuracy by +22.1\%.

\begin{table}[t!]
\centering
\small
\caption{Model performance on positive emotions and negative emotions. We report Accuracy (ACC, $\uparrow$) and VAD Distance ($D_{\text{VAD}}$, $\downarrow$). Best results are \textbf{bolded} and second best are \underline{underlined}.}
\label{tab:emotion_breakdown}
\setlength{\tabcolsep}{4pt}
\renewcommand{\arraystretch}{1.1}
\resizebox{\linewidth}{!}{
\begin{tabular}{lcccc}
\toprule
\multirow{2}{*}{Method} & \multicolumn{2}{c}{\textbf{Positive}} & \multicolumn{2}{c}{\textbf{Negative}} \\
\cmidrule(lr){2-3} \cmidrule(lr){4-5}
& ACC~$\uparrow$ & $D_{\text{VAD}}$~$\downarrow$ & ACC~$\uparrow$ & $D_{\text{VAD}}$~$\downarrow$ \\
\midrule
\multicolumn{5}{l}{\textit{\textbf{Commercial proprietary models}}} \\
Doubao-Seedream-4.0 \cite{doubao_seedream_4_0} & 66.32 & 2.00 & 56.74 & \textbf{3.09} \\
Qwen-Image-Edit-Plus \cite{qwenimagetechnicalreport} & 62.94 & 2.10 & 53.75 & 3.36 \\
Flux-kontext-pro \cite{flux} & 61.01 & 2.18 & 47.71 & 3.36 \\
Flux-kontext-max \cite{flux} & 58.94 & 2.12 & 55.00 & 3.24 \\
Doubao-Seededit-3-0 \cite{doubao_seededit_3_0} & 54.87 & 2.08 & 50.53 & 3.23 \\
\midrule
\multicolumn{5}{l}{\textit{\textbf{Open-source Models}}} \\
Step-1X \cite{step1x-edit} & 67.59 & \underline{2.00} & \underline{70.10} & 3.27 \\
Uniworld V2 \cite{uniworld} & 62.66 & 2.00 & 58.10 & 3.23 \\
OmniGen2 \cite{omnigen2} & 61.65 & 2.10 & 47.13 & 3.39 \\
Bagel \cite{bagel} & 55.39 & 2.18 & 56.61 & 3.47 \\
Flux-kontext-dev~\cite{flux}& 52.63 & 2.16 & 49.13 & 3.38 \\
DreamOmni2 \cite{dreamomni2} & 44.36 & 2.26 & 31.42 & 3.51 \\
IP2P \cite{brooks2023instructpix2pix} & 17.79 & 3.03 & 18.95 & 3.75 \\
\midrule
Qwen-Image-Edit-2509 (Base) \cite{qwenimagetechnicalreport} & \underline{73.87} & 2.02 & 50.76 & 3.11 \\
\textbf{Ours} & \textbf{75.88} & \textbf{1.96} & \textbf{72.86} & \underline{3.10} \\
\bottomrule
\end{tabular}
}
\vspace{-1.5em}
\end{table}

\subsection{Qualitative Analysis}

Figure~\ref{fig-case-study} presents visual comparisons across eight emotion editing scenarios, revealing a clear capability hierarchy among different model generations. For early models like IP2P~\cite{brooks2023instructpix2pix} and DreamOmni2~\cite{dreamomni2}, emotion editing proves fundamentally challenging. IP2P consistently outputs artifacts and semantically irrelevant changes that fail to convey target emotions, while DreamOmni2 exhibits severe editing collapse, generating outputs nearly identical to original images across multiple scenarios, indicating inability to interpret affective manipulation commands. Later models like Uniworld V2~\cite{uniworld} and FLUX-kontext-max~\cite{flux} demonstrate improved instruction understanding but rely heavily on atmospheric and color-based modifications, typically adjusting lighting, color temperature, and saturation (warm tones for positive emotions, cool desaturated palettes for negative ones) while lacking semantic depth beyond superficial visual filters. In contrast, ULMs, such as Qwen-Image-Edit+~\cite{qwenimagetechnicalreport} and Doubao-Seedream-4.0~\cite{doubao_seedream_4_0}, exhibit significantly stronger emotion editing capabilities through comprehensive visual manipulations. For instance, they convey Amusement through playful object additions and natural expression changes, and Sadness via compositional adjustments combined with appropriate color grading, demonstrating genuine multimodal understanding that bridges language instructions and visual emotional expression.

%% file: Sec/6_conclusions.tex
\section{Conclusion and Discussions}
\label{sec:conclusion}

In this paper, we introduce AIM-Bench, the first comprehensive benchmark tailored for Affective Image Manipulation. Our systematic evaluation reveals a pervasive positivity bias in current models, particularly regarding negative emotion editing. To address this, we construct AIM-40k via a novel Inverse Repainting Strategy, providing balanced and high-fidelity affective supervision. Fine-tuning on this dataset establishes a new SOTA, boosting negative emotion accuracy by over 20\%. While our evaluation pipeline demonstrates strong human-machine correlation (\textit{i.e.}, Pearson's $r > 0.73$), the reliance on closed-source MLLMs remains a constraint we plan to address in future work.

\section*{Impact Statement}

This work advances affective image manipulation (AIM) by proposing a new benchmark (AIM-Bench) and a balanced dataset (AIM-40k). These contributions facilitate applications in human-computer interaction, art therapy, and emotional content creation. However, precise emotional editing carries the risk of altering the semantic reality of visual media, potentially leading to the spread of manipulated narratives or misinformation. By addressing the inherent positivity bias in current models via targeted data, we strive for more neutral and controllable AI, but we emphasize the importance of responsible deployment and the continued development of forensic tools to detect varying degrees of affective tampering.